\documentclass[journal=jacsat,manuscript=article]{achemso}
\usepackage[version=3]{mhchem} 
\usepackage{siunitx}
\usepackage[dvipsnames]{xcolor}
\usepackage{amssymb}
\usepackage{pifont}
\usepackage{wrapfig}
\usepackage{hyperref}

\usepackage{amsmath} 

\usepackage{tabularx,colortbl}
\usepackage{capt-of}
\usepackage{adjustbox}

\author{Seongwon Kim}
\affiliation
{Department of Chemical Engineering, Carnegie Mellon University, 15213, USA}
\altaffiliation{Joint First Authorship}

\author{Parisa Mollaei}
\affiliation[meche]
{Department of Mechanical Engineering, Carnegie Mellon University, 15213, USA}
\altaffiliation{Joint First Authorship}

\author{Akshay Antony}
\affiliation[meche]
{Department of Mechanical Engineering, Carnegie Mellon University, 15213, USA}

\author{Rishikesh Magar}
\affiliation[meche]
{Department of Mechanical Engineering, Carnegie Mellon University, 15213, USA}

\author{Amir Barati Farimani}
\email{barati@cmu.edu}
\affiliation[meche]
{Department of Mechanical Engineering, Carnegie Mellon University, 15213, USA}
\alsoaffiliation[biomed]
{Department of Biomedical Engineering, Carnegie Mellon University, 15213, USA}
\alsoaffiliation[mld]
{Machine Learning Department, Carnegie Mellon University, 15213, USA}

\title[An \textsf{achemso} demo]
{GPCR-BERT: Interpreting Sequential Design of G Protein Coupled Receptors Using Protein Language Models}

\abbreviations{IR,NMR,UV}

\begin{document}

\begin{abstract}

\noindent With the rise of Transformers and Large Language Models (LLMs) in Chemistry and Biology, new avenues for the design and understanding of therapeutics have opened up to the scientific community. Protein sequences can be modeled as language and can take advantage of recent advances in LLMs, specifically with the abundance of our access to the protein sequence datasets. In this paper, we developed the GPCR-BERT model for understanding the sequential design of G Protein-Coupled Receptors (GPCRs). GPCRs are the target of over one-third of FDA-approved pharmaceuticals. However, there is a lack of comprehensive understanding regarding the relationship between amino acid sequence, ligand selectivity, and conformational motifs (such as NPxxY, CWxP, E/DRY). By utilizing the pre-trained protein model (Prot-Bert) and fine-tuning with prediction tasks of variations in the motifs, we were able to shed light on several relationships between residues in the binding pocket and some of the conserved motifs. To achieve this, we took advantage of attention weights, and hidden states of the model that are interpreted to extract the extent of contributions of amino acids in dictating the type of masked ones. The fine-tuned models demonstrated high accuracy in predicting hidden residues within the motifs. In addition, the analysis of embedding was performed over 3D structures to elucidate the higher-order interactions within the conformations of the receptors.

\end{abstract}

\section{Introduction}

Understanding the fundamental function of proteins, as the most important molecules of life, enables the development of improved therapeutics for diseases. Proteins are comprised of amino acids arranged in particular orders (sequences) that define the structures and functions of living organisms\cite{berman2002protein, branden2012introduction, lesk2010introduction, anfinsen1973principles, creighton1990protein}. The protein sequence, structure, and function are the results of millions of years of evolution and refinement driven by the survival needs of living cells\cite{patthy2009protein, wood1999evolution, golding1998structural}. One of the most fundamental questions in biology is how thousands of proteins have evolved with versatile functions with only 20 different types of amino acids as their building blocks\cite{altschul1997gapped, sadowski2009sequence, zhang2012structure}. While significant progress has been made in identifying protein structures and functions through experimental methods, bio-informatics approaches provide valuable insights by leveraging computational tools and algorithms\cite{radivojac2013large, chou2001prediction, chou2011some, tuncbag2011prediction}. For example, bio-informatics protein sequence models take advantage of co-evolution and a vast amount of proteomics and genomics data to enable prediction of structure or function. However, deciphering the complex relationships between elements of sequences (amino acids) and their functions still remains a challenge as bio-informatics models are not sophisticated enough to model higher-order interactions within amino acid sequences. In recent years, Artificial Intelligence (AI) has made significant contributions to the protein structure and function predictive models\cite{bengio2012unsupervised, lecun2015deep, goodfellow2016deep, molnar2020interpretable, radford2019language, devlin2018bert, vaswani2017attention}. Alpha-fold is a fantastic example of AI success in protein structure prediction\cite{jumper2021highly, senior2020improved, nussinov2022alphafold}. However, there has been less focus on the structure-agnostic AI bioinformatic models which predict the higher-order interactions between amino acids using sequences. With the rise of Large Language Models (LLM) and Transformers, there has been a renaissance in bio-informatics since protein sequences can be considered as language\cite{ofer2021language, madani2023large, li2021bioseq}. LLMs can take advantage of Transformer architecture and learn higher-order relations in texts\cite{devlin2018bert, brown2020language, guntuboina2023peptidebert}. In this study, we leveraged the recent advances of LLMs in protein modeling and developed a model for interpreting the G Protein Coupled Receptors (GPCRs) sequences. We developed GPCR-BERT which investigates the high-order interactions in GPCRs. The reason we chose GPCRs is that they are an essential class of cell membrane proteins found in organisms ranging from bacteria to humans\cite{strader1995family, kroeze2003g, Fredriksson, rosenbaum2009structure,lagerstrom2008structural,tuteja2009signaling, yadav2022prediction, mollaei2023activity, mollaei2023machine}. Currently, over one-third of U.S. Food and Drug Administration (FDA) approved medications target the receptors\cite{hauser2017trends, Zhang,lagerstrom2008structural, Hollingsworth_Dror, lappano2011g, Oprea, sriram2018g}. GPCRs exhibit remarkable functional diversity, including sensory perception, neurotransmission, immune response, cardiovascular regulation, vision, metabolism, and energy homeostasis\cite{lagerstrom2008structural, hauser2017trends, pierce2002seven, danielm, shukla2016g, wettschureck2005mammalian, kostenis2005g, smrcka2008g}. Although each of the receptors has its own amino acid sequence, there are a few conserved motifs among them, such as NPxxY, CWxP and E/DRY. Each conserved motif serves as a crucial component in the GPCRs\cite{wess1997g, ballesteros199519, milligan2006heterotrimeric} and the type of each x can vary according to the receptor class (varies between E and D for the E/DRY motif). We posed critical questions regarding the sequence design of GPCRs including: 1. What is the correlation between variations (xs) in the conserved region of GPCRs (as xx in NPxxY motif) and other amino acids in the sequence? 2. Can we predict the whole sequence of amino acids in a GPCR, given partial sequence? 3. Can we find amino acids having the most contributions to others that may play key roles in GPCR conformational changes? To respond to these questions, we focused on three conserved motifs (NPxxY, CWxP, and E/DRY) in different types of receptors. We defined tasks to predict the x in these motifs given the rest of the sequence and their correlations to other amino acids within the GPCR sequences using protein language models. The emergence of Transformers has further enabled machien learning (ML) models to effectively capture long-range dependencies between words and gain a deeper comprehension of language syntax\cite{vaswani2017attention, devlin2018bert}. In addition, the attention mechanism enables the model to weigh the importance of each element based on its contextual relationships within the sequence\cite{vaswani2017attention}. Prot-Bert, a pre-trained variant of Bidirectional Encoder Representations from Transformers (BERT), was employed for this research and fine-tuned with tokenized amino acid sequences. This approach facilitated the model's learning of the intrinsic patterns present within the GPCR sequence\cite{devlin2018bert,elnaggar2022prottrans}. Furthermore, we interpreted attention weights and hidden states of the fine-tuned model to extract the extent of contributions of other amino acids in determining the type of masked x amino acids in the conserved motifs. Investigating the extent of contributions of amino acids in dictating the type of x may assist us in identifying the amino acids correlated to the function of the conserved sequences. The correlated amino acids can serve as potential candidates for mutation studies and aid in generating new protein structures in protein engineering.

\section{Methods}

\begin{figure}[t!]
    \centering
    \includegraphics[width=\linewidth]{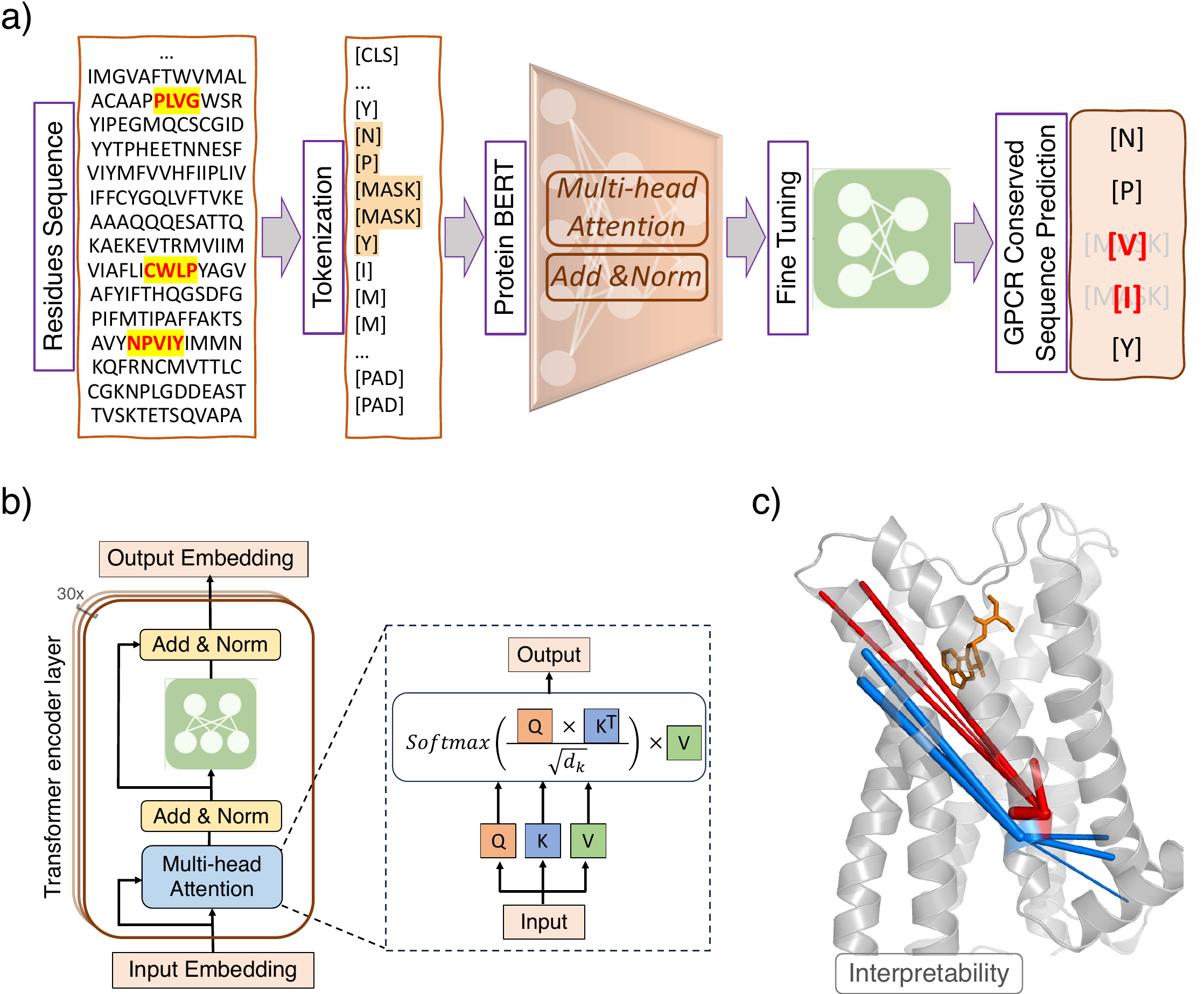}
    \caption{(a) The overall architecture of GPCR-BERT. GPCR amino acid sequences are tokenized and subsequently processed through Prot-Bert, followed by the regression head. (b) The structure of Prot-Bert Transformer and the attention layer. The input token embedding is transformed into keys, queries, and values which subsequently form the attention matrix. The output is passed through a fully connected neural network. This sequence of operations is iterated 30 times to reach the final output embedding of the GPCR sequence. (c) Representation of the top five most correlated amino acids to the first x (red) and second x (blue) in NPxxY motif within a GPCR obtained through attention heads. The thickness of the lines represents the strength of correlations (weights).}
    \label{fig:model}
\end{figure}

The overall architecture of GPCR-BERT is displayed in Fig.\ref{fig:model}(a). Prot-Bert\cite{elnaggar2022prottrans}, a transformer-based model of 16-head and 30-layer structure has been used as the pre-trained model. Prot-Bert is pre-trained on a massive protein sequence corpus, UniRef100\cite{uniref}, which contains over 217 million unique sequences. A variant of the original BERT\cite{devlin2018bert} developed by Google, this architecture is derived from the encoder segment of the Transformer model, comprised of multiple, sequential layers of an attention-feed-forward network (Fig.\ref{fig:model}(b)). Within the attention structure, each token is encoded into input embedding which is converted into keys, queries, and values. Keys and queries are combined via matrix multiplication to form the attention map, which is subsequently passed through a softmax\cite{Bridle_1990} function to generate a probability distribution. Following this process, the resulting distribution is used to scale (multiply) the value vectors. The feed-forward layer within each Transformer layer facilitates the learning of intricate patterns embedded in the input, whereas the attention mechanism is responsible for understanding and encoding the relationships among various tokens. The multi-head attention structure divides the input across multiple parallel attention layers, or "heads". This setup enables each head to independently learn and specialize in capturing different types of patterns and relationships\cite{vaswani2017attention}. This structure of the Transformer encoder in Prot-Bert enables the model to learn context-aware representations of amino acids in a protein sequence by considering each sequence as a "document". The model has been trained with a Masked Language Model (MLM) objective, which is a training method where some percentage of input tokens are masked at random, and the model must predict those masked tokens from their context. One of the biggest advantages of ProT-Bert over some other pre-trained models (such as Prot-XL and Prot-Albert) is that its performance has exceeded these models on several benchmarks\cite{elnaggar2022prottrans}. For fine-tuning process, a series of experiments were performed to identify the most suitable architecture for the regression head. As a result, the design of three fully connected layers is selected.

\subsection{Data preprocessing}

The GPCR sequences dataset used for fine-tuning Prot-Bert was obtained from the GPCRdb database \cite{horn2003gpcrdb}. The GPCRdb database serves as a comprehensive resource of valuable experimental data on G-protein coupled receptors (GPCRs). This includes details on GPCR binding, configurations, and signal transduction. Moreover, the database offers a range of analysis tools and computational data, such as homology models and multiple sequence alignments, which further enhance the understanding and exploration of GPCRs\cite{horn2003gpcrdb, rose2016rcsb, berman2003announcing}. From the GPCRdb, all class A GPCR sequences that contain either NPxxY, CWxP or E/DRY motifs are extracted, obtaining a dataset of 293 protein sequences. Among the extracted protein sequences, some of them were incomplete sequences embodying missing residues that are denoted as 'X' which were especially prevalent in relatively long sequences. The aim was to remove the ones that contain excessive missing residues and avoid inefficient padding in the tokenization process. Considering the distribution of sequence lengths (Fig.S1), we knocked out the sequences that are larger than 370 which is 13.3\% of initial data. This filtering process led to a dataset of 254 class A GPCR sequences distributed across 62 different receptor classes (Table S1). Although the three motifs are 'highly' conserved, not every single class A GPCR contains these motifs. For example, residue C in CWxP motif is only conserved 71\% among class A GPCRs\cite{Olivella2013cysteine6.47}.  Thus, the initial dataset from GPCRdb was refined to 3 separate datasets to include only those sequences that embody each conserved motif. This process yielded a dataset composed of 238 sequences for the NPxxY prediction task, 168 for the CWxP task and 212 for the E/DRY task. 

\noindent In the tokenization process, we searched through each sequence for the location of the conserved motifs ensuring their proper localization within specific regions: NPxxY in Transmembrane (TM) 7, CWxP in TM6, and E/DRY in TM3.  The 'x' residue earmarked for prediction (for example, xx in the NPxxY) was substituted with 'J', an alphabet character absent from the Prot-Bert tokenizer vocabulary. Simultaneously, label sequences were assembled, each retaining the ground truth amino acid while substituting all other residues with 'J'. During tokenization, the 'J' characters in the input sequences were replaced with the [MASK] token. Utilizing the Prot-Bert tokenizer, a [CLS] token was inserted at the beginning of each sequence, while sequences falling short of the maximum input length were padded using the [PAD] token. Both the input and label sequences were then translated into corresponding integers as dictated by the tokenizer vocabulary.

\section{Results}

\subsection{Prediction tasks of conserved motifs}

\begin{table}[t!]
    \begin{adjustbox}{width=\columnwidth,center}
    \begin{tabular}{ |c|c|c|c|c|c| } 
    \hline
    Conserved Motif & Number of data & Train Loss & Test Loss & Train Accuracy & Test Accuracy \\
    \hline
    NPxxY & 238 & 0.138 $\pm$ 0.020 & 0.213 $\pm$ 0.015 & 99.25 $\pm$ 0.323 & 98.05 $\pm$ 0.479\\
    CWxP& 168& 0.376 $\pm$ 0.040& 0.702 $\pm$ 0.040& 89.94 $\pm$ 0.468& 86.29 $\pm$ 1.010\\
    E/DRY& 212& 0.092 $\pm$ 0.0006& 0.089 $\pm$ 0.0006& 100& 100 \\
    \hline
    \end{tabular}
    \end{adjustbox}
    \caption{\label{tab:acc1}Fine-tuning result of each prediction task. The outcomes pertaining to loss and accuracy for each downstream task are displayed.  Each task is averaged over the results of three runs. }
\end{table}

\begin{figure}[t!]
    \centering
    \includegraphics[width=1\linewidth]{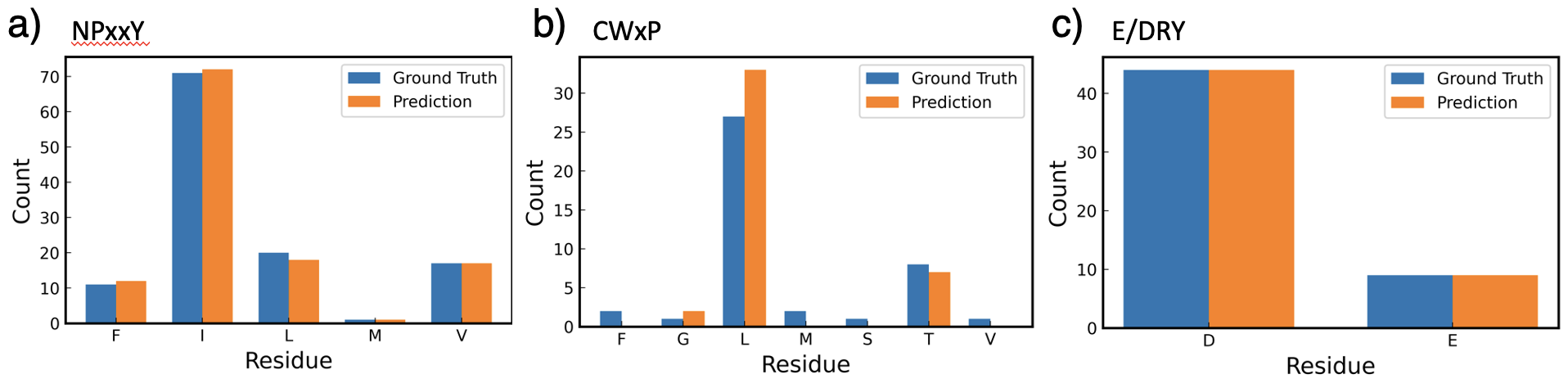}
    \caption{The test result for the downstream tasks of predicting x in (a) NPxxY, (b) CWxP, and (c) E/DRY. Each task was subjected to training for 30 epochs prior to testing. For the NPxxY task, the prediction results for both masked positions were taken into account.}
    \label{fig:barplot}
\end{figure}

The curated datasets were partitioned into training and testing subsets to a ratio of 0.75:0.25, and were subsequently incorporated into the structure of Prot-Bert, complemented by a regression head. A series of experiments were performed to identify the suitable architecture for the regression head. The optimal architecture was chosen as the three fully connected layers of multi-layer perceptrons with respective node counts of 1024, 256, and 30. This specific configuration yielded the highest prediction accuracy. Additionally, the model incorporated a dropout rate of 0.25 and the Rectified Linear Unit (ReLU)\cite{Nair1990relu} activation function. An Adam optimizer\cite{Diederik1990adam} was employed alongside a learning rate (LR) scheduler to facilitate the model training. The LR scheduler was designed to diminish the learning rate by a factor of 0.2, contingent on stagnation or lack of improvement in the loss function. The loss function chosen for this task was Cross Entropy Loss, which represents a composite of a softmax activation function and a negative log-likelihood loss function. This choice was particularly appropriate for the multi-class prediction task at hand, as it imposes more substantial penalties on the model when erroneous predictions are made with high confidence. Furthermore, multi-class accuracy was assessed by contrasting the output of the regression head with the label tokens. The model was trained on a single NVIDIA GeForce GTX 2080 Ti GPU with 12GB of memory. The results of the fine-tuning process are delineated in Table\ref{tab:acc1} and Fig.\ref{fig:barplot}. The NPxxY task and E/DRY task demonstrated exceptional performance, achieving nearly perfect accuracy with minimal error. In comparison, the performance on the CWxP task was lower, yet within an acceptable range of accuracy. This relative disparity in performance, particularly for the CWxP task, is assumed to be attributed to the smaller dataset size compared to the other two tasks.

\subsection{Prediction tasks of masked sequences}

\begin{table}[t!]
    \begin{adjustbox}{width=0.8\columnwidth,center}
    \begin{tabular}{| c|c|c|c|c |} 
    \hline
    Task& Train Loss & Test Loss & Train Accuracy & Test Accuracy \\
    \hline
    5 masked & 0.174 $\pm$ 0.015 & 0.837 $\pm$ 0.040 & 99.26 $\pm$ 0.184 & 83.07 $\pm$ 0.661\\
    10 masked & 0.159 $\pm$ 0.026 & 1.193 $\pm$ 0.025 & 99.32 $\pm$ 0.127 & 74.57 $\pm$ 1.502\\
    15 masked & 0.154 $\pm$ 0.021 & 1.166 $\pm$ 0.015 & 99.31 $\pm$ 0.336 & 76.72 $\pm$ 0.440\\
    50 masked & 0.134 $\pm$ 0.010 & 1.254 $\pm$ 0.056 & 99.58 $\pm$ 0.233 & 74.89 $\pm$ 1.115\\
    \hline
    \end{tabular}
    \end{adjustbox}
    \caption{\label{tab:acc2}Result of masked sequence prediction task. Each task is averaged over the results of three runs. }
\end{table}

To evaluate GPCR-BERT's capability to predict other parts of the sequence than the conserved motifs and to find out whether it is possible to infer a complete GPCR sequence from partial information, we have tested the model with several masked sequence prediction tasks as shown in Table\ref{tab:acc2}. The dataset containing all 254 GPCR sequences is used since these tasks are not related to separate conserved motifs. For each GPCR sequence, residues for each task are continuously masked with 'J' tokens starting from the sequence index 100 which is a randomly selected number. The number of masking is increased to 10, 15 and 50 to further examine the model's prediction capability. The 5 masked prediction task shows the highest prediction accuracy of 83.07\% while the task of 10, 15, 50 shows similar accuracy of around 75\%. The decent prediction accuracy indicates that GPCR-BERT is capable of predicting large number of residues and not just a small number of residues in the conserved motifs.

\subsection{Comparison with other models}

\begin{table}[t!]
    \begin{adjustbox}{width=0.8\columnwidth,center}
    \begin{tabular}{| c|c|c|c|c|} 
    \hline
    Model & E/DRY & NPxxY & CWxP & 5 Masked\\ 
    \hline
    GPCR-BERT & 100 & 98.05 $\pm$ 0.479 & 86.29 $\pm$ 1.010 & 83.07 $\pm$ 0.661\\ 
    BERT & 83.02 & 59.85 $\pm$ 0.266 & 71.43 & 10.97 $\pm$ 0.312\\
    SVM & 90.57 & 76.67 & 71.43 & 63.49\\ 
    \hline
    \end{tabular}
    \end{adjustbox}
    \caption{\label{tab:acc3}Test accuracy (\%) of each prediction task of different ML models. Each task is averaged over the results of three runs.}
\end{table}

To compare the GPCR-BERT's performance with other ML models, we tested several prediction tasks to the original BERT and the Support Vector Machine (SVM). Compared with BERT, we can investigate the effect of leveraging an encoder pre-trained on protein sequences as opposed to one trained on generic English corpus. By testing on SVM, we can investigate how low-key ML models can perform when trained with limited protein sequence data. SVM is a popular supervised learning algorithm that is known for its high accuracy, ability to handle high-dimensional data, and versatility in modeling both linear and non-linear relationships. SVM works by finding a hyperplane that best divides the dataset into classes. The main objective of the SVM is to maximize the margin between data points and the separating hyperplane. The best hyperplane is the one for which this margin is maximized\cite{Noble2006svm}. 

\noindent The dataset used for all models have the same train/test split ratio and random state. As shown in the result(Table\ref{tab:acc3}), BERT was able to predict quite well (83\% accuracy) in a relatively simple E/DRY task but the accuracy decreased significantly for more intricate tasks like NPxxY and the 5 masked prediction. From this, we can confirm the advantage of utilizing pre-trained models on protein database corpus in the interpretation of protein sequences. Interestingly, although the performance was incomparable to the GPCR-BERT, SVM has shown better results than the original BERT. This suggests that BERT's pre-training on English text may have adversely affected the learning of protein sequences.

\section{Discussion}
\subsection{Interpretability}

The attention mechanism of the Transformer model facilitates the capturing of global sequence information within each token embedding in the encoder. Yet, in practical applications, the classification token, known as the [CLS] token, often serves as the aggregate representation of the entire sequence\cite{schwaller1}\cite{schwaller2} The BERT tokenizer introduces the [CLS] token at the commencement of every sequence, which is then employed by BERT to assign the sequence to specific classes. Thus, the [CLS] token encapsulates information from all output embedding. Using this insight, we aimed to verify whether GPCR-BERT effectively recognizes and differentiates among various GPCR types. To this end, we extracted the [CLS] tokens of each GPCR sequence from the final hidden state which has the structure of 1024 nodes and visualized with t-distributed Stochastic Neighbor Embedding (t-SNE)\cite{Maaten1990tsne}. T-SNE executes dimensionality reduction by determining pairwise similarities in the high-dimensional space and constructing a Gaussian joint probability distribution. Correspondingly, a similar probability distribution is defined using a Student's t-distribution. The algorithm then seeks to minimize the Kullback-Leibler divergence between these two distributions. This minimization process inherently attracts similar data points toward each other while repelling dissimilar points, thereby enabling the clustering of similar data points. The results are shown in Fig.\ref{fig:tsne_weight}(a). 
Interpreting the attention weights allows us to examine how the GPCR-BERT is able to understand the grammar of GPCR sequences. Within the hierarchical structure of multiple attention layers in the encoder, the terminal layer is most likely to capture advanced semantic information in the context of Natural Language Processing tasks, and analogous structural or binding information for protein sequences\cite{devlin2018bert}\cite{bertology}. Thus, the 16 attention heads of the last layer of GPCR-BERT are extracted and visualized through heatmaps (Fig.\ref{fig:tsne_weight}(b)). High attention weights are indicative of a greater degree of correlation between the corresponding residues in the input sequence. These elevated attention weights highlight specific regions of the sequence that the model perceives as containing valuable information. Attention allows every token in the input sequence to have a direct connection with all other tokens in the sequence, thereby facilitating the modeling of inter-dependencies between tokens irrespective of their relative positions\cite{vaswani2017attention}. In addition, the softmax function of the attention enables the attention weight to serve as a probability distribution resulting in the sum of each row being 1. Thus, the attention weights of the GPCR-BERT represent the relative correlation of the corresponding residue with others. To scrutinize the interrelationships between conserved motifs and the remaining sequence in GPCRs, the highest five weights associated with variant residues within the conserved motifs are examined. In the context of the CWxP prediction downstream task, for instance, the attention weight corresponding to 'x' is selected, and the top five values are analyzed for every GPCR sequence. The five residues displaying the highest correlation are further investigated using PyMOL\cite{pymol} to examine 3D crystal structure and the biological relevance of GPCR-BERT's interpretation of GPCRs. The findings of this analysis are presented in Fig.\ref{fig:corr_res}, Fig.\ref{fig:prot_seq}.

\subsection{Mapping Higher Order Interactions}

 \begin{figure}[t!]
     \centering
     \includegraphics[width=1.0\linewidth]{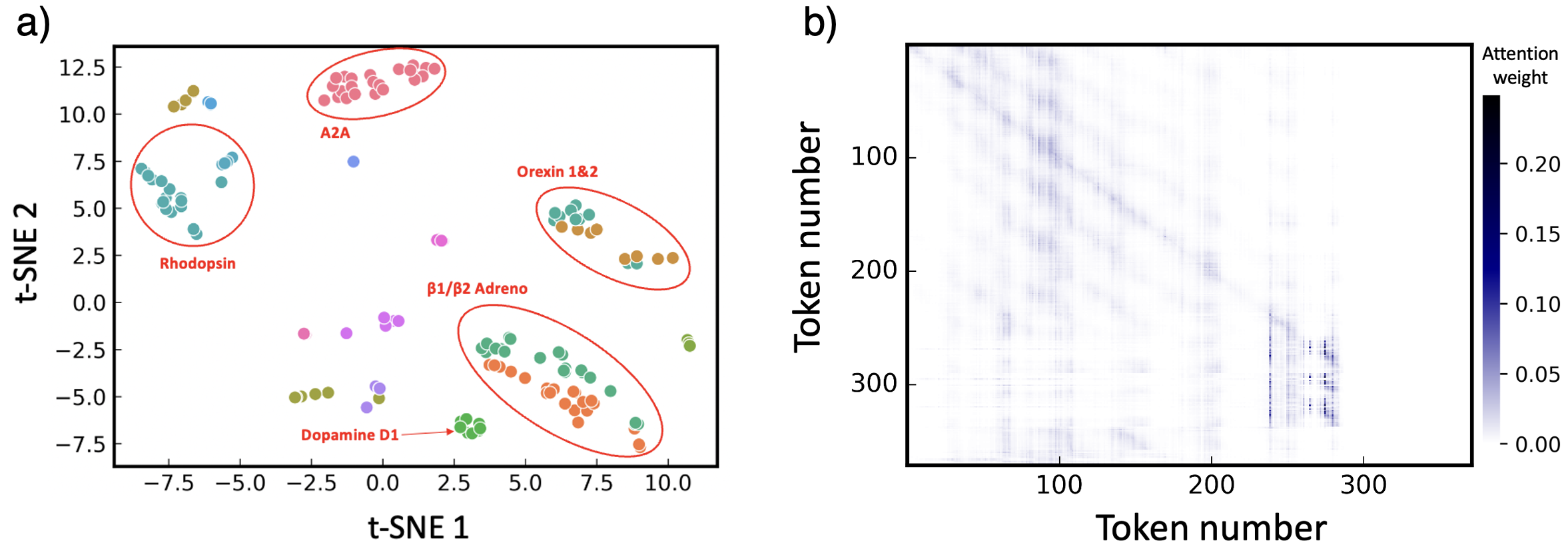}
     \caption{(a) t-SNE visualization of GPCRs. t-SNE was applied to the [CLS] token embedding from the last hidden state of GPCR-BERT to reduce the dimension to 2D. Different receptor classes are denoted by distinct colors and the clustering of the same GPCR type is obvious. (b) Attention weight heatmap of human $\beta$ 2 adreno receptor GPCR (4GBR). The figure provides a visual representation of head 1 of the final attention layer in GPCR-BERT. The labels on the axes correspond to the respective tokens.}
     \label{fig:tsne_weight}
 \end{figure}
\begin{table}[t!]
    \begin{adjustbox}{width=0.85\columnwidth,center}
    \begin{tabular}{|c|c|c|c|l|}
    \hline
    \textbf{Head} & \textbf{Repetition}& \textbf{Residue} & \textbf{Location (BW)} & \multicolumn{1}{c|}{\textbf{Matching with Mutagenesis data}}     \\ \hline
    2             & 16                & D in DFRIAF      & 8.49 (H8)              & no match                                        \\ \hline
    4             & 18                & L in LPFFIV      & 6.49                   & no match                                        \\ \hline
    4             & 15                & G in GIIMGT      & 6.38                   & no match                                        \\ \hline
    4             & 17                & H in HVIQDN      & 6.58                   & no match                                        \\ \hline
    5             & 9                 & K in KFERLQ      & 1.59                   & no match                                        \\ \hline
    7             & 11                & T in TKTWTF      & 2.66                   & \textbf{M 2.66 (receptor-expression)}           \\ \hline
    7             & 17                & I in IQMHWY      & 4.61                   & no match                                        \\ \hline
    11            & 11                & K in KTWTFG      & 2.67                   & \textbf{1 before M 23.49 (receptor-expression)} \\ \hline
    11            & 19                & E in EFWTSI      & 3.26                   & no match                                        \\ \hline
    11            & 12                & K in KEHKAL      & 6.28                   & \textbf{2 after C 6.27 (thermostabilization)}\\ \hline
    11            & 9                 & E in EAKRQL      & 5.64                   & no match                                        \\ \hline
    11            & 11                & E in ERLQTV      & 12.48 (ICL 1)          & no match                                        \\ \hline
    12            & 6                 & I in ILTKTW      & 2.64                   & \textbf{1 after H 2.63 (receptor-ligand-bond)}  \\ \hline
    12            & 14                & G in GAAHIL      & 2.6                    & \textbf{3 before H 2.63 (receptor-ligand-bond)} \\ \hline
    14            & 9                 & C in CLKEHK      & 6.27                   & \textbf{C 6.27}                                 \\ \hline
    14            & 7                 & K in KFCLKE      & 6.25                   & \textbf{2 before C 6.27 (receptor-ligand-bond)} \\ \hline
    14            & 6                 & L in LAVVPF      & 2.55                   & no match                                        \\ \hline
    15            & 20                & R in RQLQKI      & 5.67                   & no match                                        \\ \hline
    15            & 19                & R in RVFQEA      & 5.6                    & no match                                        \\ \hline
    15            & 6                 & K in KIDKSE      & 5.71                   & no match                                        \\ \hline
    16            & 17                & H in HVIQDN      & 6.58                   & no match                                        \\ \hline
    16            & 7                 & P in PFGAAH      & 2.58                   & no match                                        \\ \hline
\end{tabular}
    \end{adjustbox}
    \caption{\label{tab:acc4}Comparison of GPCR-BERT correlation results with the mutagenesis data in GPCRdb for $\beta$ 2 Adreno receptors.  Columns labeled 'head' and 'repeated' display the respective head number that detected the correlation and the frequency of its occurrence. 'Residue' and 'position' columns represent the residue and its position as per the Ballesteros-Weinstein (BW) numbering system. The 'Match with mutagenesis data' column indicates if the identified correlation aligns with experimental mutation results.}
\end{table}

\begin{figure}[t!]
     \centering
     \includegraphics[width=0.95\linewidth]{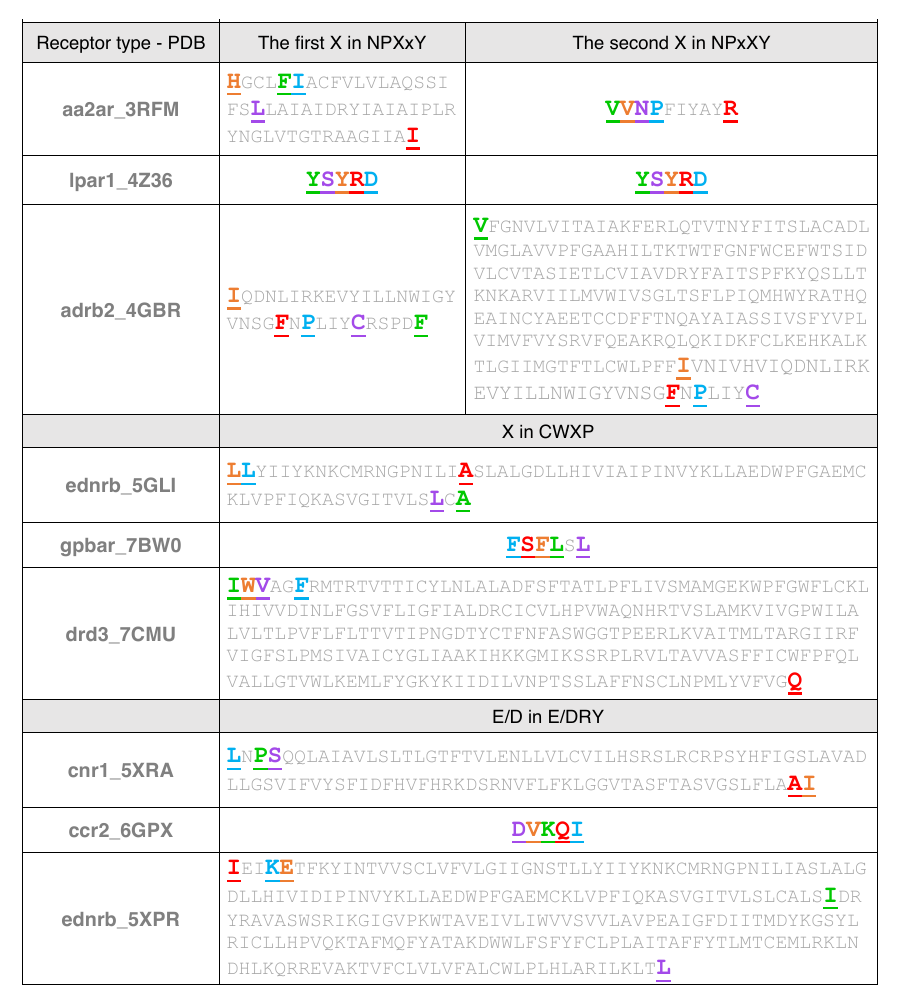}
     \caption{Top five (red (first), orange (second), blue (third), green (fourth), and purple (fifth)) most correlated amino acids to the x amino acids within NPxxY, CWxP, and E/DRY motifs in different GPCRs.}
     \label{fig:corr_res}
 \end{figure}
 
The fine-tuned GPCR-BERT's representations of GPCRs are depicted via a t-SNE plot (Fig.\ref{fig:tsne_weight}(a)). For the sake of effective visualization, only classes consisting of more than three GPCRs were selected. As evidenced by the t-SNE plot, GPCRs of the same receptor class (depicted by identical color markers) are clustered together, thereby indicating the model's successful classification of GPCRs based solely on their sequence information. This result also suggests that the [CLS] token within the embedding effectively captures the distinguishing characteristics of the individual GPCR classes. The model's ability to differentiate between classes such as the Orexin1 and Orexin2 receptor and the $\beta$1/$\beta$2 Adreno receptor - both of which possess the NPIIY motif - implies that the model takes into account the entire GPCR sequence, rather than merely the conserved motifs when executing prediction tasks. Moreover, given that the inter-cluster distances within the t-SNE plot signify the degree of similarity between each cluster, it is evident that the GPCR-BERT model is able to discern the subtle variations of GPCR classes. This capability is particularly notable in the case of distinguishing between $\beta$1 and $\beta$2 Adreno receptor classes depicted in the lower right corner of Fig.\ref{fig:tsne_weight}(a). More detailed version of this plot can be found in Fig.S2 (see Supporting Information). Fig.\ref{fig:tsne_weight}(b) illustrates the attention maps of the tokenized GPCR (4GBR) sequence as processed by the last layer of GPCR-BERT. The observed pattern of attention weight distribution remained consistent across the GPCR receptor class (refer to Fig.S3 and Fig.S4 in Supporting Information for heatmap of all 16 heads), suggesting that GPCR-BERT was successful in learning representations that underscore the interrelations of tokens within the sequence. As shown in Fig.\ref{fig:tsne_weight}(b), the padding region (far right) received an attention weight of 0, while a noticeably strong attention weight was detected visualized in dark colors. This indicates that GPCR-BERT identified these residues as being more significantly correlated than others. This pattern was consistently observed across several heads (2, 3, 6, 8, 12, 16) of 4GBR, and extended to other receptor classes of GPCRs. Notably, this pattern was particularly prevalent in heads 1 and 2, where the attention weights exceeded 0.1 in the corresponding region. Therefore, an analysis of these patterns is expected to yield valuable insights into the conserved motifs.

\begin{figure}[t!]
     \centering
     \includegraphics[width=0.82\linewidth]{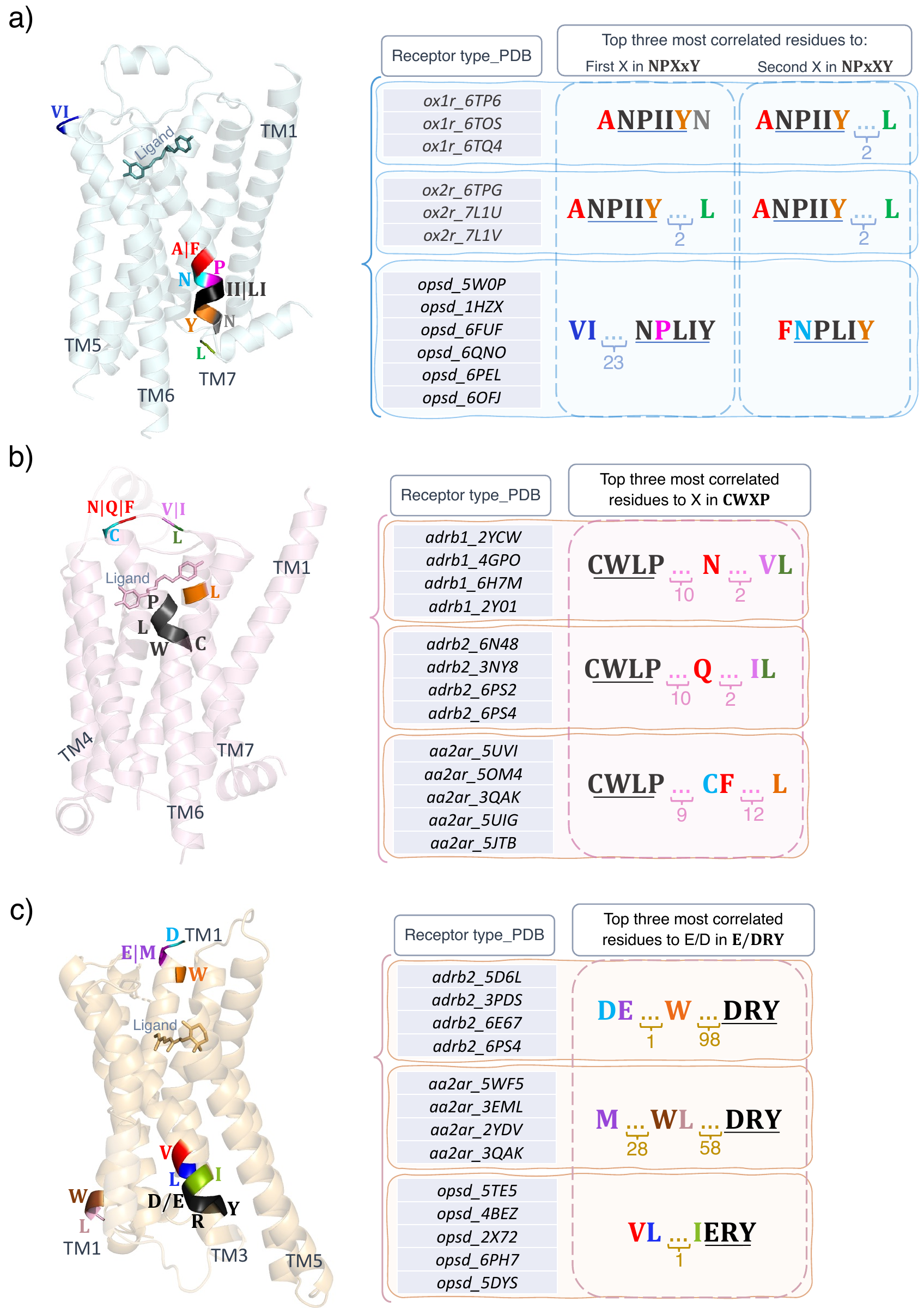}
     \caption{Structural representations for the top three most correlated amino acids to (a) xx in NPxxY, (b) x in CWxP, and (c) E/D in E/DRY motifs within various GPCRs. In the tables, the colors of the correlated amino acids correspond to the colors assigned to amino acids across the protein and the numbers denote the number of residues separating them.}
     \label{fig:prot_seq}
 \end{figure}

\noindent Fig.\ref{fig:corr_res} introduces the top five most correlated amino acids to the x residues in the NPxxY and CWxP motifs as well as the E/D residue in E/DRY motif for a few receptors in the dataset (see Data availability for the comprehensive list of GPCRs and their correlated residues). The common pattern found for attention and consequently correlation between NPxxY motif and other residues is that the xx residues are mainly attending to their adjacent amino acids (NP). The first x in NPxxY usually attends to the upper top region of the receptors in the extracellular segment of protein. This higher-order interaction possibly demonstrates the connection between NPxxY motif and the ligand-selective region of the receptor.\cite{farimani2018binding} This is significant because the xx residues vary between different types of GPCRs. Our finding elucidates why changes in the GPCR type correlate to xx residues in NPxxY. Fig.\ref{fig:prot_seq} shows structural representations for the top three most correlated amino acids to the xs in NPxxY (a) and CWxP (b) motifs and E/D in E/DRY motif within various receptors. The correlated amino acids are displayed in colors and the numbers indicate the number of amino acids separating them. As the figure shows, similar types of receptors are categorized based on the correlated amino acid predictions. As noted earlier, the xx in NPxxY motif mostly attend to neighboring residues and some residues at the extracellular part of the receptor that are responsible for ligand selectivity. However, the x in CWxP motif and E/D in E/DRY motif mainly attend to some residues in the binding pocket. We can conclude that the message passed through the ligand in the binding pocket may be modulated by the variations in x and E/D residues within the NPxxY, CWxP, and E/DRY motifs. This finding will help us to focus on the sequential design of these motifs in selecting the ligand. 

\noindent Furthermore, we investigated whether residues with high correlation, other than those adjacent to the conserved motifs, align with experimental mutation findings. Thus, our correlation findings were compared with the mutagenesis data in the GPCRdb\cite{horn2003gpcrdb}. This data contains the accumulated results of the positions of mutated residues and their effects on GPCRs. Table\ref{tab:acc4} shows the results for the comparison for $\beta$ 2 adreno receptor (results for other classes are shown in Table S2\~S4). It can be observed that head 6 and 10 of the GPCR-BERT found the residues that are related to receptor expression significant while head 11 and 13 discerned the significance of residues linked to thermostabilization. We can also see that multi-head attention mechanism of transformer models enables each head to learn distinct patterns and this characteristic can facilitate protein sequence analysis. Additionally, some highly correlated residues, such as residue E 3.26 in head 10, were identified by GPCR-BERT as potentially significant but remain understudied. These insights may provide a foundation for further investigations into GPCR structures and their mechanisms. There are, however, limitations to this study considering the minor inconsistency of the conserved motifs in class A GPCRs. As mentioned earlier, residue C in CWxP motif is only conserved 71\% \cite{Olivella2013cysteine6.47} while the NPxxY motif is conserved 94\% among class A GPCRs. The remaining variations are in NPxxxY (3\%) or NPxxF (3\%) form\cite{Oliveira1999npxxf}. Similarly, Y in the E/DRY motif has exceptions as well\cite{Tao2014dry}. Since the model is focused on predicting only the general type of these motifs, its understanding of GPCRs might not apply to ones that have varied versions of the motifs.

\section{Conclusion}

In this study, we explored the higher-order interactions in sequence design of G protein-coupled receptors (GPCRs) and their impact on determining the type of function by taking advantage of language models. We utilized the Prot-Bert model, a transformer-based language model, and fine-tuned it with tokenized amino acid sequences to predict non-conserved amino acids (depicted by x) in the conserved motifs of NPxxY, CWxP, and E/DRY in various GPCRs. The results showed that the model effectively predicted these residues in the receptors with high accuracy. The attention weights and hidden states of the model were also analyzed to understand the extent of contributions of other amino acids in dictating the type of masked ones. We demonstrated that GPCR-BERT successfully distinguished different GPCR classes based on their sequences, and has the potential for understanding functional relationships within protein structures. These findings could also have implications in mutation studies, protein engineering, drug development, and further advancing our understanding of GPCR biology. 

\section{Data and software availability}
The necessary information containing the codes and data for downstream tasks used in this study are available here: 
\href{https://github.com/Andrewkimmm/GPCR-BERT}{https://github.com/Andrewkimmm/GPCR-BERT}

\section{Supporting Information}
The Supporting Information includes a complete figure of the t-SNE of GPCR-BERT and heatmaps visualizing all 16 attention heads of several GPCRs.

\begin{acknowledgement}

This work is supported by the Center for Machine Learning in Health (CMLH) at Carnegie Mellon University and a start-up fund from Mechanical Engineering Department at CMU. 

\end{acknowledgement}

\bibliography{reference}

\end{document}


\maketitle

\section{Data distribution of initial dataset}

\begin{figure}[t!]
    \centering
    \includegraphics[width=0.95\linewidth]{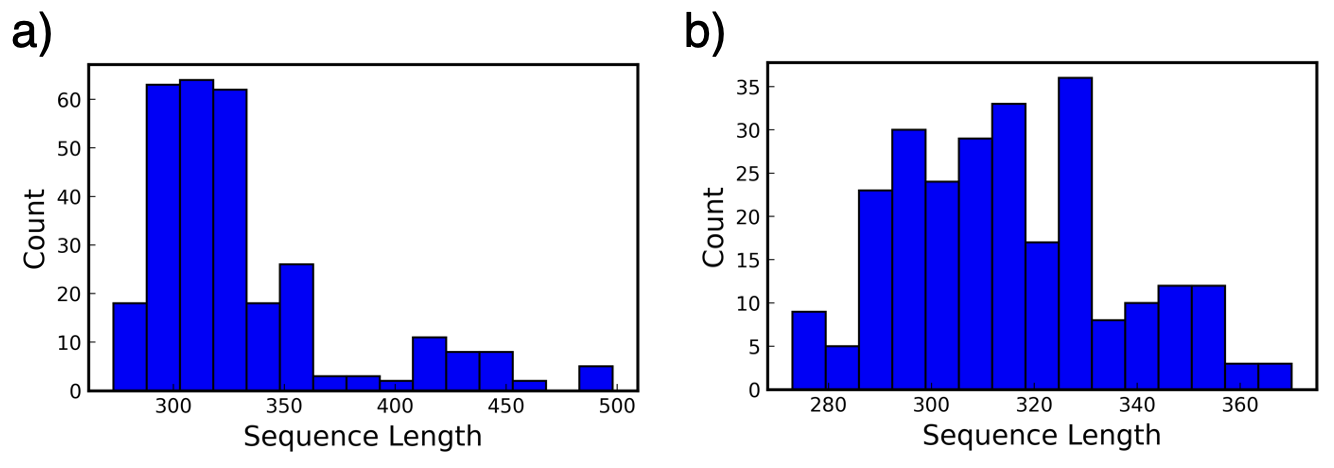}
    \caption{Distribution histogram of the sequence length of GPCRs in the initial dataset (a) and after filtering (b). }
    \label{fig:data_hist}
\end{figure}

Fig.S\ref{fig:data_hist} displays the distribution of GPCR data sequence lengths that are initially extracted from GPCRdb and after filtering process. 293 Class A GPCRs that embody either NPxxY, CWxP or E/DRY motifs are extracted from the GPCRdb (Fig.S\ref{fig:data_hist}(a)). To avoid sequences with excessive missing residues and ensure efficient tokenization, we excluded sequences exceeding 370 residues leading to a filtered dataset of 254 sequences (Fig.S\ref{fig:data_hist}(b)). This dataset consists of 62 receptor classes and distribution is displayed in Table S\ref{tab:S1}. 

\begin{table}[t!]
    \begin{adjustbox}{width=0.65\columnwidth,center}
    \begin{tabular}{| c|c|l|l|l|l|l|l|} 
    \hline
    aa2ar& 24&       drd1&    9&cnr1&5&5ht2a&4\\
    adrb1& 23&       ox2r&8&cnr2&4&mtr1a&4\\ 
    adrb2& 21&       5ht2b&8&ntr1&4&ebnrb&4\\ 
 opsd& 20&       nk1r&6&cltr2&4& others&83\\
 ox1r& 14&       cxcr4&5&gpr52&4&total&254\\
 \hline 
    \end{tabular}
    \end{adjustbox}
    \caption{\label{tab:S1} Receptor class distribution of dataset, only classes that have more than 3 data are displayed. The dataset consists of 62 receptor classes with 44 of them having less than 4 data.}
\end{table}

\section{Fully Labeled t-SNE}

\begin{figure}[t!]
    \centering
    \includegraphics[width=0.9\linewidth]{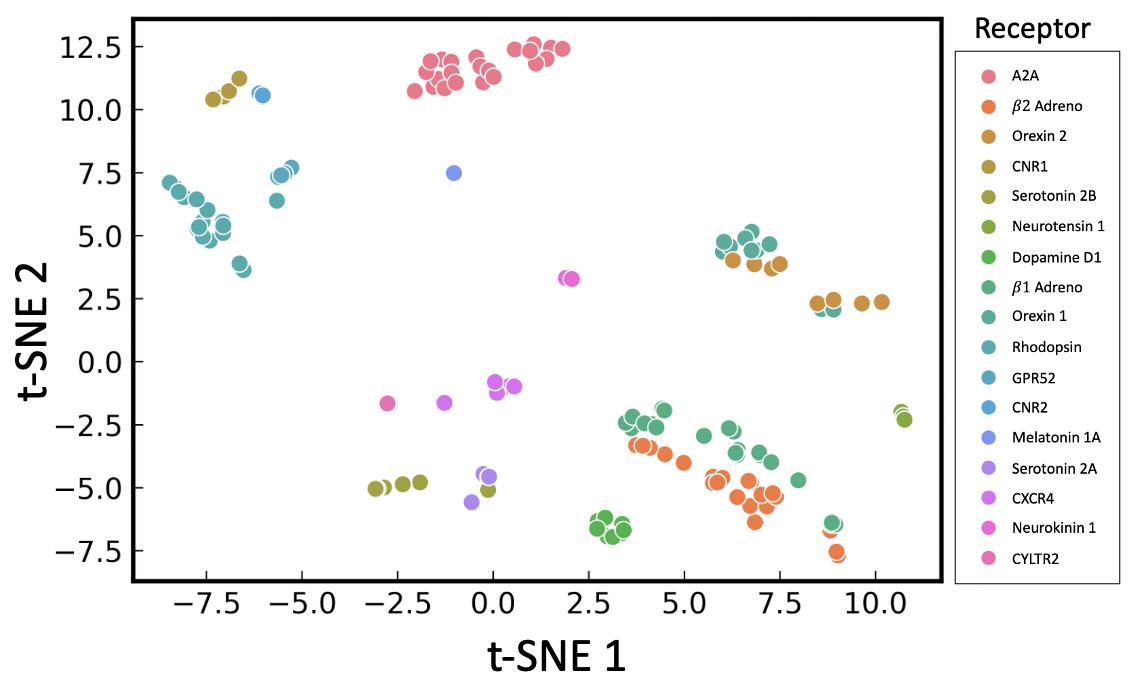}
    \caption{Fully labeled t-SNE of GPCRs. The [CLS] token embedding from the last hidden state has been subjected to both dimension reduction and clustering techniques. }
    \label{fig:full-tsne}
\end{figure}

Fig.S\ref{fig:full-tsne} displays the fully labeled t-SNE plot from GPCR-Bert with distinct colors representing different receptor classes. [CLS] token embeddings are extracted from the last hidden state and visualized. Note that from the complete dataset, only classes comprising more than three data points have been taken into account. Notably, even classes with a small number of data points are clearly distinguished. For example, GPCR-Bert effectively recognizes the subtle sequential difference between CNR1 and CNR2 receptor classes (upper left corner).

\section{Attention head heatmap}

Heatmaps in Fig.S\ref{fig:beta-attention} and Fig.S\ref{fig:dopa-attention} present a graphical illustration of the last attention layer in GPCR-Bert. Attention weights of all 16 heads are extracted and visualized, with darker colors representing larger weights. The multi-head attention mechanism allows the model to capture distinct patterns of different parts of the input embeddings as each head is assigned to investigate different parts.  A comparative analysis of the heatmaps reveals that GPCR-Bert gave a similar interpretation of sequences according to their classes comparing Fig.S\ref{fig:beta-attention} ($\beta$ adreno receptor class) with Fig.S\ref{fig:dopa-attention} (Dopamine 1 receptor class). In addition, several heads such as Head 1 and Head 2 displays similar pattern irrespective of different GPCR classes.

\begin{figure}[t!]
    \centering
    \includegraphics[width=0.95\linewidth]{figs/2attention(beta).png}
    \caption{Attention weight heat map of (a) $\beta$ 2 adreno receptor GPCR(4GBR), (b) $\beta$ 1 adreno receptor GPCR(4AMI). The figure provides a visual representation of the final attention layer. }
    \label{fig:beta-attention}
\end{figure}

\begin{figure}[t!]
    \centering
    \includegraphics[width=0.95\linewidth]{figs/2attention(dopa).png}
    \caption{Attention weight heat map of (a) Dopamine 1 receptor GPCR(7LJC) and (b) Dopamine 1 receptor GPCR(7CKX). The figure provides a visual representation of the final attention layer.}
    \label{fig:dopa-attention}
\end{figure}

\section{Comparison with Mutagenesis data}

Table S\ref{tab:S2} to S\ref{tab:S4} shows the comparison with the correlation results of the GPCR-BERT and the mutagenesis data from the GPCRdb. The top 5 residues with the highest attention weights were identified from the final layer of GPCR-BERT and compared with the experimental mutagenesis results in GPCRdb. From the results, we can observe that some portion of the residues that GPCR-BERT identified as significant residues align with the experimental mutation studies.
\begin{table}[t!]
    \begin{adjustbox}{width=0.75\columnwidth,center}
    \begin{tabular}{|c|c|c|c|l|}
    \hline
    \textbf{Head} & \textbf{Repetition} & \textbf{Residue} & \textbf{BW}       & \textbf{Matching with Mutagenesis data}        \\ \hline
    4             & 4                 & K in KGIIAI      & 4.43              & \textbf{K 4.43 (thermostabilization)}          \\ \hline
    4             & 4                 & K in KSAAII      & 6.35              & \textbf{2 before L 6.37 (thermostabilization)} \\ \hline
    4             & 24                & H in HAAKSL      & 6.32              & \textbf{5 before L 6.37 (thermostabilization)} \\ \hline
    4             & 7                 & K in KGIIAI      & 4.43              & \textbf{K 4.43 (thermostabilization)}          \\ \hline
    5             & 20                & S in SLLAIA      & 3.42              & \textbf{3 after S 3.39 (thermostabilization)}  \\ \hline
    6             & 24                & F in FCPDCS      & 6.60              & \textbf{no match}                              \\ \hline
    7             & 19                & T in TLQKEV      & 6.25              & no match                                       \\ \hline
    8             & 9                 & A in APLWLM      & 7.30              & \textbf{no match}                              \\ \hline
    9             & 24                & H in HGCLFI      & 3.23              & no match                                       \\ \hline
    9             & 24                & I in IACFVL      & 3.28              & \textbf{no match}                              \\ \hline
    9             & 12                & F in FIACFV      & 3.27              & no match                                       \\ \hline
    10            & 24                & H in HAAKSA      & 6.32              & \textbf{5 before L 6.37 (thermostabilization)} \\ \hline
    10            & 20                & K in KEVHAA      & 6.29              & \textbf{no match}                              \\ \hline
    11            & 8                 & K in KGIIA       & 4.43              & \textbf{K 4.43 (thermostabilization)}          \\ \hline
    11            & 16                & R in RAKGII      & 4.41              & \textbf{2 before K 4.43 (thermostabilization)} \\ \hline
    11            & 18                & Q in QKEVHA      & 6.28              & \textbf{no match}                              \\ \hline
    12            & 24                & F in FVVSLA      & 2.42              & \textbf{4 before L 2.46 (thermostabilization)} \\ \hline
    12            & 24                & Y in YFVVSL      & 2.41              & \textbf{5 before L 2.46 (thermostabilization)} \\ \hline
    13            & 24                & T in TGTRAA      & 4.38              & \textbf{5 before K 4.43 (thermostabilization)} \\ \hline
    13            & 22                & T in TGTRAK      & 4.36              & no match                                       \\ \hline
    13            & 18                & A in ACLFED      & 4.69 (End of TM4) & no match                                       \\ \hline
    13            & 20                & A in APLWL       & 7.30              & no match                                       \\ \hline
    14            & 24                & L in LRYNGL      & 34.51 (ICL2)      & no match                                       \\ \hline
    15            & 8                 & H in HGCLFI      & 3.23              & no match                                       \\ \hline
    15            & 17                & L in LAARRQ      & 5.63              & \textbf{L 5.63 (thermostabilization)}          \\ \hline
    16            & 23                & I in IPFAIT      & 2.57              & \textbf{6 before T 2.62 (thermostabilization)} \\ \hline
    16            & 7                 & VVSLAA           & 2.44              & \textbf{2 before L 2.46 (thermostabilization)} \\ \hline
    \end{tabular}
    \end{adjustbox}
    \caption{\label{tab:S2} Comparison of GPCR-BERT correlation results with the mutagenesis data in GPCRdb for A2a receptors.  Columns labeled 'head' and 'repeated' display the respective head number that detected the correlation and the frequency of its occurrence. 'Residue' and 'position' columns represent the residue and its position as per the Ballesteros-Weinstein (BW) numbering system. The 'Match with mutagenesis data' column indicates if the identified correlation aligns with experimental mutation results.}
\end{table}

\begin{table}[t!]
    \begin{adjustbox}{width=0.75\columnwidth,center}
    \begin{tabular}{|c|c|c|c|c|}
    \hline
    \textbf{Head} & \textbf{Repetition} & \textbf{Residue} & \textbf{BW}    & \textbf{Matching with Mutagenesis data}        \\ \hline
    1             & 8                   & R in RYQSL       & 34.52 (ICL 52) & no match                                       \\ \hline
    1             & 15                  & T in TLVVRG      & 2.62           & no match                                       \\ \hline
    3             & 6                   & K in KEQIRK      & 5.56           & no match                                       \\ \hline
    3             & 6                   & D in DFRKAF      & 8.49 (H8)      & no match                                       \\ \hline
    3             & 8                   & Y in YCRSPD      & 7.53           & \textbf{Y 7.53 (thermostabilization)}          \\ \hline
    4             & 18                  & C in CRSPDF      & 7.54           & \textbf{1 after Y 7.53 (thermostabilization)}  \\ \hline
    6             & 15                  & F in FKRLLA      & 8.54 (H8)      & \textbf{5 before C 8.59 (palmitoylation-site)} \\ \hline
    7             & 6                   & I in IGSTQR      & 1.57           & \textbf{2 before R 1.59 (thermostabilization)} \\ \hline
    10            & 9                   & E in EQIRKI      & 5.67           & no match                                       \\ \hline
    10            & 7                   & R in RKIDR       & 5.70           & no match                                       \\ \hline
    11            & 9                   & Q in QIRKID      & 5.68           & no match                                       \\ \hline
    11            & 9                   & E in EQIRKI      & 5.67           & no match                                       \\ \hline
    11            & 17                  & Q in QRLQTL      & 12.48 (ICL 1)  & \textbf{2 after R 1.59 (thermostabilization)}\\ \hline
    12            & 15                  & R in RYQSLM      & 34.52 (ICL 2)  & no match                                       \\ \hline
    12            & 11                  & K in KVIICT      & 4.43           & no match                                       \\ \hline
    12            & 9                   & R in RARAKV      & 4.39           & no match                                       \\ \hline
    12            & 14                  & Q in QIRKID      & 5.68           & no match                                       \\ \hline
    13            & 13                  & V in VWAISA      & 4.49           & no match                                       \\ \hline
    13            & 14                  & T in TSPFRY      & 3.55           & no match                                       \\ \hline
    13            & 17                  & T in TNRAYA      & 5.34           & no match                                       \\ \hline
    13            & 10                  & D in DFRKAF      & 8.49 (H8)      & no match                                       \\ \hline
    14            & 9                   & K in KVIICT      & 4.43           & no match                                       \\ \hline
    14            & 6                   & N in NLFITS      & 2.40           & no match                                       \\ \hline
    15            & 12                  & R in RKIDR       & 5.70           & no match                                       \\ \hline
    15            & 9                   & R in RAKVII      & 4.41           & no match                                       \\ \hline
    15            & 7                   & R in RSPDFR      & 7.55           & no match                                       \\ \hline
    16            & 9                   & P in PFGAT       & 2.58           & \textbf{5 after M 2.53 (thermostabilization)}  \\ \hline
    \end{tabular}
    \end{adjustbox}
    \caption{\label{tab:S3} Comparison of GPCR-BERT correlation results with the mutagenesis data in GPCRdb for $\beta$ 1 Adreno receptors.  Columns labeled 'head' and 'repeated' display the respective head number that detected the correlation and the frequency of its occurrence. 'Residue' and 'position' columns represent the residue and its position as per the Ballesteros-Weinstein (BW) numbering system. The 'Match with mutagenesis data' column indicates if the identified correlation aligns with experimental mutation results.}
\end{table}

\begin{table}[t!]
    \begin{adjustbox}{width=0.75\columnwidth,center}
    \begin{tabular}{|c|c|c|c|c|}
    \hline
    \textbf{Head} & \textbf{Repetition} & \textbf{Residue} & \textbf{BW}   & \textbf{Matching with Mutagenesis data}         \\ \hline
    1             & 11                  & E in ESFVIY      & 5.36          & no match                                        \\ \hline
    1             & 13                  & L in LLIMLG      & 1.41          & no match                                        \\ \hline
    2             & 18                  & Q in QFRNCM      & 8.49 (H8)     & no match                                        \\ \hline
    2             & 13                  & C in CMVTTL      & 8.53 (H8)     & no match                                        \\ \hline
    3             & 18                  & Q in QFRNCM      & 8.49 (H8)     & \textbf{no match}                               \\ \hline
    3             & 16                  & C in CMVTTL      & 8.53 (H8)     & \textbf{no match}                               \\ \hline
    4             & 13                  & G in GGFTTT      & 2.55          & \textbf{1 before G 2.56 (INCL disease causing)} \\ \hline
    4             & 13                  & G in GFTTTL      & 2.56          & \textbf{G 2.56 (INCL disease causing)}          \\ \hline
    4             & 18                  & G in GGEIAL      & 3.35          & no match                                        \\ \hline
    4             & 13                  & V in VVCKPM      & 3.53          & no match                                        \\ \hline
    4             & 15                  & G in GQLVFT      & 5.59          & no match                                        \\ \hline
    5             & 18                  & L in LYVTVQ      & 4.54          & no match                                        \\ \hline
    5             & 17                  & G in GFTTTL      & 2.56          & \textbf{G 2.56 (INCL disease causing)}          \\ \hline
    5             & 18                  & F in FTTTLY      & 2.57          & \textbf{1 after G 2.56 (INCL disease causing)}  \\ \hline
    6             & 15                  & L in LICWLP      & 6.45          & \textbf{5 after M 6.40 (receptor expression)}   \\ \hline
    6             & 15                  & C in CMVTTL      & 8.53 (H8)     & no match                                        \\ \hline
    7             & 14                  & T in TTTLYT      & 2.56          & \textbf{2 after G 2.56 (INCL disease causing)}  \\ \hline
    7             & 18                  & V in VVCKPM      & 3.53          & no match                                        \\ \hline
    7             & 18                  & V in VVVCKP      & 3.52          & no match                                        \\ \hline
    7             & 18                  & Y in YVVVCK      & 3.51          & no match                                        \\ \hline
    7             & 18                  & T in TLYVTV      & 1.53          & no match                                        \\ \hline
    8             & 15                  & A in APPLVG      & 4.58          & no match                                        \\ \hline
    8             & 18                  & G in GFPINF      & 1.46          & no match                                        \\ \hline
    8             & 18                  & A in AFTWVM      & 4.47          & no match                                        \\ \hline
    8             & 11                  & A in AIERYV      & 2.47          & no match                                        \\ \hline
    8             & 14                  & S in SNFRFG      & 34.52         & no match                                        \\ \hline
    9             & 14                  & E in EGFFAT      & 3.28          & \textbf{E 3.28 (receptor expression)}\\ \hline
    9             & 18                  & L in LGGEIA      & 3.34          & no match                                        \\ \hline
    9             & 12                  & E in EIALWS      & 3.37          & no match                                        \\ \hline
    9             & 13                  & Q in QFRNCM      & 8.49 (H8)     & no match                                        \\ \hline
    10            & 18                  & F in FCYGQL      & 5.56          & no match                                        \\ \hline
    10            & 18                  & R in RYVVVC      & 3.50          & no match                                        \\ \hline
    10            & 17                  & G in GFTTTL      & 2.56          & \textbf{G 2.56 (INCL disease causing)}          \\ \hline
    10            & 18                  & F in FTTTLY      & 2.57          & \textbf{1 after G 2.56 (INCL disease causing)}  \\ \hline
    10            & 18                  & G in GGFTTT      & 2.55          & \textbf{1 before G 2.56 (INCL disease causing)} \\ \hline
    10            & 9                   & F in FMVFGG      & 2.52          & no match                                        \\ \hline
    11            & 18                  & K in KLRTPL      & 12.49 (ICL 1) & no match                                        \\ \hline
    11            & 14                  & T in TTTLYT      & 2.56          & \textbf{2 after G 2.56 (INCL disease causing)}  \\ \hline
    11            & 18                  & L in LNYILL      & 2.39          & no match                                        \\ \hline
    11            & 16                  & I in ILLNLA      & 2.42          & no match                                        \\ \hline
    11            & 17                  & A in AIERYV      & 2.47          & no match                                        \\ \hline
    12            & 18                  & N in NYILLN      & 2.40          & no match                                        \\ \hline
    12            & 18                  & L in LYVTVQ      & 1.54          & no match                                        \\ \hline
    12            & 13                  & L in LLNLAV      & 2.43          & no match                                        \\ \hline
    12            & 18                  & N in NYILLN      & 2.40          & no match                                        \\ \hline
    13            & 18                  & A in AEPWQF      & 1.27          & no match                                        \\ \hline
    13            & 15                  & I in IMLGFP      & 1.43          & no match                                        \\ \hline
    13            & 11                  & T in TLGGEI      & 3.33          & no match                                        \\ \hline
    13            & 11                  & V in VCKPMS      & 3.54          & no match                                        \\ \hline
    \end{tabular}
    \end{adjustbox}
    \caption{\label{tab:S4} Comparison of GPCR-BERT correlation results with the mutagenesis data in GPCRdb for Rhodopsin receptors.  Columns labeled 'head' and 'repeated' display the respective head number that detected the correlation and the frequency of its occurrence. 'Residue' and 'position' columns represent the residue and its position as per the Ballesteros-Weinstein (BW) numbering system. The 'Match with mutagenesis data' column indicates if the identified correlation aligns with experimental mutation results.}
\end{table}